# Multi-Modality Conditioned Variational U-Net for Field-of-View Extension in Brain Diffusion MRI

Zhiyuan Li*[a], Chenyu Gao,[a] Praitayini Kanakaraj,[b] Shunxing Bao,[a] Lianrui Zuo,[a] Michael E. Kim,[b] Nancy R. Newlin,[b] Gaurav Rudravaram,[a] Nazirah M. Khairi,[a] Yuankai Huo,[a,b] Kurt G. Schilling,[c] Walter A. Kukull,[d] Arthur W. Toga,[e] Derek B. Archer,[f,g] Timothy J. Hohman,[f,g] Bennett A. Landman,[a,b,c]

[a]Department of Electrical and Computer Engineering, Vanderbilt University, Nashville, TN, USA
[b]Department of Computer Science, Vanderbilt University, Nashville, TN, USA
[c]Department of Radiology & Radiological Sciences, Vanderbilt University Medical Center, Nashville, TN, USA
[d]Department of Epidemiology, University of Washington, Seattle, Washington, USA
[e]Laboratory of Neuro Imaging, Stevens Neuroimaging and Informatics Institute, Keck School of Medicine, University of Southern California, Los Angeles, CA, USA
[f]Vanderbilt Memory & Alzheimer's Center, Vanderbilt University Medical Center, Nashville, TN USA
[g]Vanderbilt Genetics Institute, Vanderbilt University Medical Center, Nashville, TN USA

**Abstract**

An incomplete field-of-view (FOV) in diffusion magnetic resonance imaging (dMRI) can severely hinder the volumetric and bundle analyses of whole-brain white matter connectivity. Although existing works have investigated imputing the missing regions using deep generative models, it remains unclear how to specifically utilize additional information from paired multi-modality data and whether this can enhance the imputation quality and be useful for downstream tractography. To fill this gap, we propose a novel framework for imputing dMRI scans in the incomplete part of the FOV by integrating the learned diffusion features in the acquired part of the FOV to the complete brain anatomical structure. We hypothesize that by this design the proposed framework can enhance the imputation performance of the dMRI scans and therefore be useful for repairing whole-brain tractography in corrupted dMRI scans with incomplete FOV. We tested our framework on two cohorts from different sites with a total of 96 subjects and compared it with a baseline imputation method that treats the information from T1w and dMRI scans equally. The proposed framework achieved significant improvements in imputation performance, as demonstrated by angular correlation coefficient (p < 1E-5), and in downstream tractography accuracy, as demonstrated by Dice score (p < 0.01). Results suggest that the proposed framework improved imputation performance in dMRI scans by specifically utilizing additional information from paired multi-modality data, compared with the baseline method. The imputation achieved by the proposed framework enhances whole brain tractography, and therefore reduces the uncertainty when analyzing bundles associated with neurodegenerative.

*Corresponding Author, E-mail: zhiyuan.li@vanderbilt.edu

## 1 Introduction

Diffusion magnetic resonance imaging (dMRI) provides an in-vivo and non-invasive approach that measures the displacement of water molecules in biological tissues and has become an essential technique for studying the microstructure and connectivity of brain white matter [1–

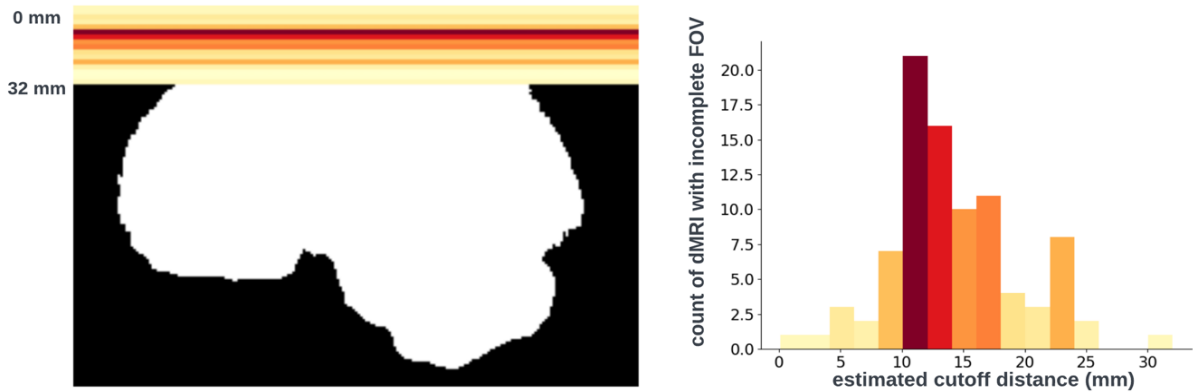

**Figure 1. Visualization (left) and histogram (right) of 103 real cases of dMRI scans with incomplete FOV that failed quality assurance. In the left figure, horizontal regions indicate the distribution of the incomplete part of FOV with an estimated position of a brain mask. The total cutoff distance from the incomplete FOV to the top of the brain is estimated using a corresponding and registered T1w image. Its histogram is presented in the right figure.**

4]. Typically, dMRI scans require multiple 3D volumes for capturing water self-diffusion in different diffusion-weighting directions, thereby producing four-dimensional-space (4D) diffusion-weighted images (DWI). For each three-dimensional-space (3D) volume, a specific direction of diffusion-encoding magnetic gradient pulses will be applied to reflect the tissue properties that restrict the movement of water molecules. The less constrained the self-diffusion of water molecules is along the applied gradient direction, the larger the observed signal attenuation will be. This gradient direction is commonly reported as a unit-length vector known as the b-vector. The amount of attenuation is characterized by a variable called the b-value, which encapsulates the duration and strength of the gradient pulse. Reference volumes with no signal attenuation are labeled as b0 images, and are required for dMRI scans to provide a baseline signal intensity[5]. Various metrics, like mean diffusivity (MD) and fractional anisotropy (FA), are quantifiable metrics calculated from these multi-volume DWIs using a diffusion tensor model to provide a simplified representation of directional water self-diffusion in brain tissues [6]. Additionally, downstream analysis like white matter tractography can be conducted to map white matter pathways in the brain, offering insights into the whole-brain connections [7,8]. Over the past ten years, dMRI and its associated analysis have become preferred approaches for investigating brain tissue characteristics for Alzheimer's Disease, stroke, schizophrenia, and the effects of general aging [7,9–14].

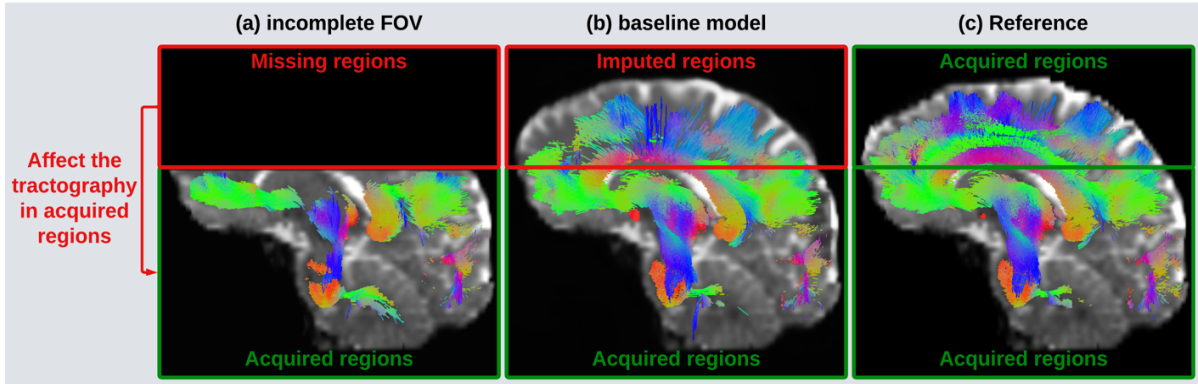

**Figure 2.** An incomplete FOV not only makes it impossible to analyze the missing regions but can also impact the tractography performed in the acquired regions (panel (a)), e.g., resulting in missing streamlines of various tracts (e.g., CST) compared with reference (panel (c)). Although existing works can somewhat impute the missing regions, they encounter difficulty when imputing slices close to the top edge of the brain, which therefore affects the tractography in those regions (panel (b)).

One major challenge of dMRI in practice is the long acquisition time needed to capture multiple volumes at various gradient directions. dMRI protocols typically suggest using more than 31 directions for detailed studies on disease progression or treatment effects [15]. This extended acquisition time can further introduce and amplify imaging artifacts, such as patient motion and eddy-current induced distortions [5,16,17], and consequently resulting in dMRI scans that have incomplete FOV, which is one of the most common problems detected during dMRI data quality checks [18]. In our recent analysis of dMRI datasets [19], 103 out of 1057 cases exhibited incomplete FOV due to such challenges (Figure 1). The absence of data from these missing regions not only hinders analyses in the incomplete part of the FOV but also severely impacts the accuracy of tractography across the entire brain, including brain regions that lie inside the FOV (Figure 2 (a)). These corrupted data then leave gaps among the complete analysis of patients and can introduce obstacles in diagnosing and monitoring neurological conditions like Alzheimer's Disease [19–21].

Traditional imputation methods for dMRI fail to address the FOV extension tasks due to the requirement of multi-volumes reference signals [19], which are unfortunately unavailable in the incomplete part of the FOV across all volumes. Recently, deep learning methods have been showing promising performance for medical image synthesis tasks including dMRI, such as distortion correction [22], denoising [23–25], and registration [26,27]. Furthermore, information from other modalities has proven to be beneficial in enhancing the quality of synthesized images. For example, T1-weighted (T1w) images are widely used for providing anatomical reference for many dMRI synthesis tasks [19,22,28,29]. Aligning with these approaches, our previous work demonstrated that a basic image-to-image translation artificial neural network, conditioned on T1-weighted (T1w) images, effectively enables the imputation of a large range of incomplete FOV in dMRI [19], and the imputation is then useful for improving the accuracy of whole brain tractography. However, this straightforward approach exhibits three primary limitations. First, we observed that this baseline method struggles to impute slices near the superior-most edges of the brain (Figure 2 (b)). Second, this baseline method treats the information from T1w images and dMRI images equally by simply concatenating them in the input layer of the neural network. This preliminary attempt raises further questions regarding the optimal use of the unique information from different modalities, and how to specifically utilize these other modalities for improving the quality of dMRI imputations. Third, other conditional and derived information for dMRI, like b-vector direction and diffusion tensor images (DTI) respectively, have not been considered as a group of multi-modality information besides T1w images. In summary, because the diffusion features in a single DWI volume are highly consistent under the same b-value and b-vector, the diffusion features observed in the acquired regions should be critical for accurately imputing the incomplete part of the FOV for the same volume. Therefore, there is a great need for advancing the way we use this available information in acquired dMRI with incomplete FOV and applying it to the anatomical structure of missing brain regions guided by information from other available modalities.

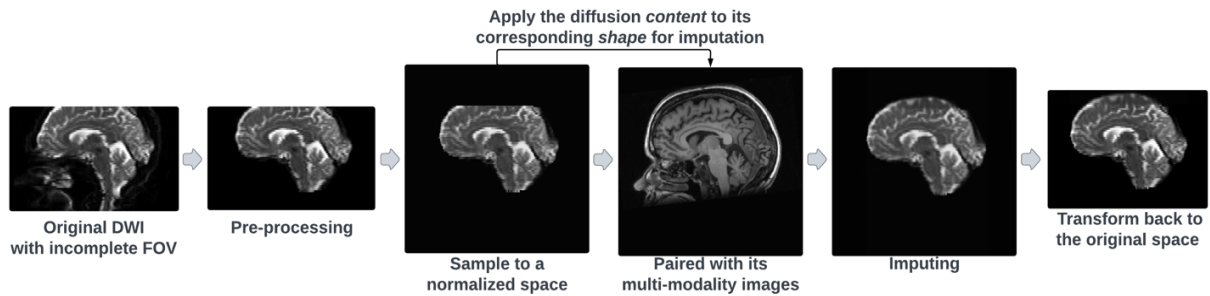

Figure 3. The pipeline of the proposed FOV extension framework begins with the preprocessing of a DWI with an incomplete FOV. Then, the DWI is sampled into a normalized space. Next, the learned diffusion features are spatially broadcast to its multi-modality shape condition to perform the imputation of the missing regions. Finally, the imputed missing regions are resampled back to their original space and concatenated with the original acquired DWI and therefore complete the FOV.

To address these limitations, we first draw inspiration from the Variational U-Net [30] which models the complex interplay between the inherent shape and appearance of images. This network then learns a generative model to synthesize images based on these characteristics. For example, it can generate images of a person wearing a specified outfit (representing appearance) while adopting a particular pose (representing shape). This concept of image synthesis aligns with the objectives of dMRI imputation. By applying the diffusion features learned from acquired dMRI to the anatomical structures contained in multi-modality images, such as T1w images, we can more accurately impute the missing brain regions of DWI. This process encourages that the synthesized regions possess the same diffusion characteristics as the acquired dMRI while maintaining consistency with the anatomical brain structure conditioned by multi-modality images. However, the Variational U-Net was not originally designed for imputation tasks and does not explicitly model how to spatially apply the features from one region to another. Especially, in dMRI imputation, we need to apply distinct characteristics from multiple volumes to the same anatomical structures. To overcome this limitation, we turn to a simple yet effective network architecture known as the spatial broadcast decoder [31]. This model tiles (broadcasts) the latent feature vector across space and concatenates it with a predefined coordinating grid. This architecture introduces a structured prior of positional information, thereby improving the disentanglement of features and the quality of synthesized images. Inspired by this approach, we plan to spatially broadcast features learned from acquired dMRI into the incomplete regions of the FOV that need to be imputed.

With the above solution ideas in mind, we propose a multi-modality conditioned variational U-net model integrated with a spatial broadcasting method to impute missing regions caused by incomplete FOV in brain dMRI. This framework tackles the imputation task with two specialized modules, each designed to learn and utilize the distinct information from various modalities. First, a *content* module learns diffusion features from the acquired regions in dMRI and spatially broadcasts this *content* to the incomplete space of FOV. Second, a *shape* module extracts features from a group of multi-modality images, including T1-weighted (T1w) images with a complete FOV and orientation maps of diffusion tensor imaging (DTI) that may have an incomplete FOV. These *shape* features then provide a prior of brain anatomical structure for imputation. A spatial broadcast decoder is adapted to integrate the *content* features with the *shape* guidance to impute the missing regions. The entire model is optimized end-to-end using stochastic gradient descent. We hypothesize that the proposed framework can enhance the imputation performance of incomplete FOV for whole volume dMRI, and this imputation can be useful for further improving the accuracy of whole brain tractography compared to our baseline method. Therefore, it can repair corrupted dMRI scans, serving as a desirable alternative to discarding the valuable data. We trained and evaluated our framework in two cohorts from different sites. The experiment results demonstrated the proposed model can be useful for improving the imputation performance in dMRI and increasing the accuracy of the downstream tractography analysis. Figure 3 illustrates the pipeline of the proposed framework for dMRI imputation.

## 2  Method

*2.1 Datasets*

We follow our previous work [19] and chose the Wisconsin Registry for Alzheimer's Prevention (WRAP) dataset [32] as our primary resource for training and evaluating our methodologies for two main reasons. First, the data from WRAP was collected at a single site, providing a consistent setting for training and evaluating models without the complications of inter-site variability. Second, the WRAP dataset provides a great opportunity to explore incomplete FOV, given that there is missing cortical slices up to 30mm in some cases. Our initial cohort for the WRAP study included 323 participants, all of whom had T1w image and single-shell dMRI scans with a b-value of 1300 s/mm², which is the most common b-value acquired in WRAP protocols. These participants were divided into three groups: 231 in the training set, 46 in the validation set, and 46 in the testing set. To further assess the robustness and generalizability of our method, we expanded our study to incorporate the National Alzheimer's Coordinating Center (NACC) dataset [33], which contains a substantial number of dMRI scans also acquired at a b-value of 1300 s/mm². Our second cohort included 50 test subjects from NACC, each with a T1w image and single-shell dMRI scans at a b-value of 1300 s/mm².

*2.2 Data preprocessing*

The PreQual [34] preprocessing pipeline was run for all DWI for denoising, solving inter-volume motion, fixing slices corrupted by signal dropout, and correcting susceptibility- and eddy-current-induced artifacts. Quality checks were conducted on the preprocessing reports and resulting images to verify valid inputs and successful data preprocessing. Both DWI and T1w images underwent intensity normalization separately. The maximum intensity value was adjusted to the 99.9th percentile of each scan, and the minimum value was set to zero. Next, all DWI are resampled and padded to isotropic 1mm space with a size of $256 \times 256 \times 256$ using MRtrix3 "mrgrid" command [35]. Then the corresponding T1w image was registered to this isotropic 1mm space DWI using FSL's "epi reg" [36] command that applies an affine transformation computed between the T1w image and the average b0 image of the DWI. Additionally, we computed the DTI of the acquired regions for each DWI using FSL's DTIFIT [37,38], where we note DTI have an identical FOV to the DWI. The angles of the estimated tensor's major eigenvector between the three main axes are computed and concatenated together as a three-channel DTI orientation image. Finally, we created b-vector map for each volume, where each b-vector map is a two-channel image, where each channel is the repeating value of the theta and phi for the polar coordinates of the b-vector. Some conditioned modality images can be shared across all volumes, such as the T1w image, and some conditioned images are unique for each volume, such as the b-vector maps. The complete view of all multi-modality conditions is presented in the panel named "all input data" in Figure 4.

*2.3 Model*

We are interested in imputing a brain DWI in the incomplete part of the FOV. Given an input of an N-volume diffusion-weighted image $X = \{x_1, x_2, ..., x_N\}$ that may have incomplete FOV, we want to learn a mapping, $G : X \rightarrow Y$, such that $Y$ will have complete imputed FOV. Because each volume $x_v$ $(v = 1,2,3,...,N)$ contains different diffusion features of the brain, we tackle the imputation task of $X$ by separately imputing each $x_v$. Specifically, for each volume we denote the acquired regions as $x_v^+$ and the missing regions that require imputation as $x_v^-$. First, we learn an inference model $q$ that can extract the diffusion features $z_v$ from $x_v^+$. Next, we prepare each $x_v$ with its corresponding set of multi-modality conditioned images $C_v = \{c_1, c_2, ..., c_M\}$ as anatomical structure guidance for imputation, where $M$ is the number of modalities. Then, we spatially apply the diffusion features $z_v$ along with the multi-modality images $C_v$ and learn a generative model $p(x_v^- | z_v, C_v)$ synthesizing the missing regions and therefore imputing the incomplete parts of the FOV. Finaly, the synthesized missing regions are then combined with the acquired regions to produce the final imputed volume, $\tilde{x}_v$. For simplicity, the sub-index $v$ is omitted in the remainder of this paper. The inference model $q$ is implemented by a neural network encoder parameterized by $\phi$, and the generative model $p$ is implemented by a U-Net-like spatial broadcast decoder parameterized by $\theta$. Both $\theta$ and $\phi$ can be optimized by minimizing the learning objective of a conditional variational autoencoder (VAE) [39] as:

$$\mathcal{L}_{VAE}(\theta, \phi) = D_{KL}[q_\phi(z|x^+, C)||p(z)] - \mathbb{E}_{q_\phi(z|x^+, C)}[\log p_\theta(x^-|z, C)], \quad (1)$$

where the first term is KL divergence between the inferred posterior distribution of the diffusion features and its prior distribution that is implemented as an isotropic Gaussian distribution parameterized as $\mathcal{N}(0, I)$, and the second term is the expectation of the negative log-likelihood of the missing regions, which is implemented as reconstruction loss of the imputed missing regions supervised by its ground truth DWI, $y$.

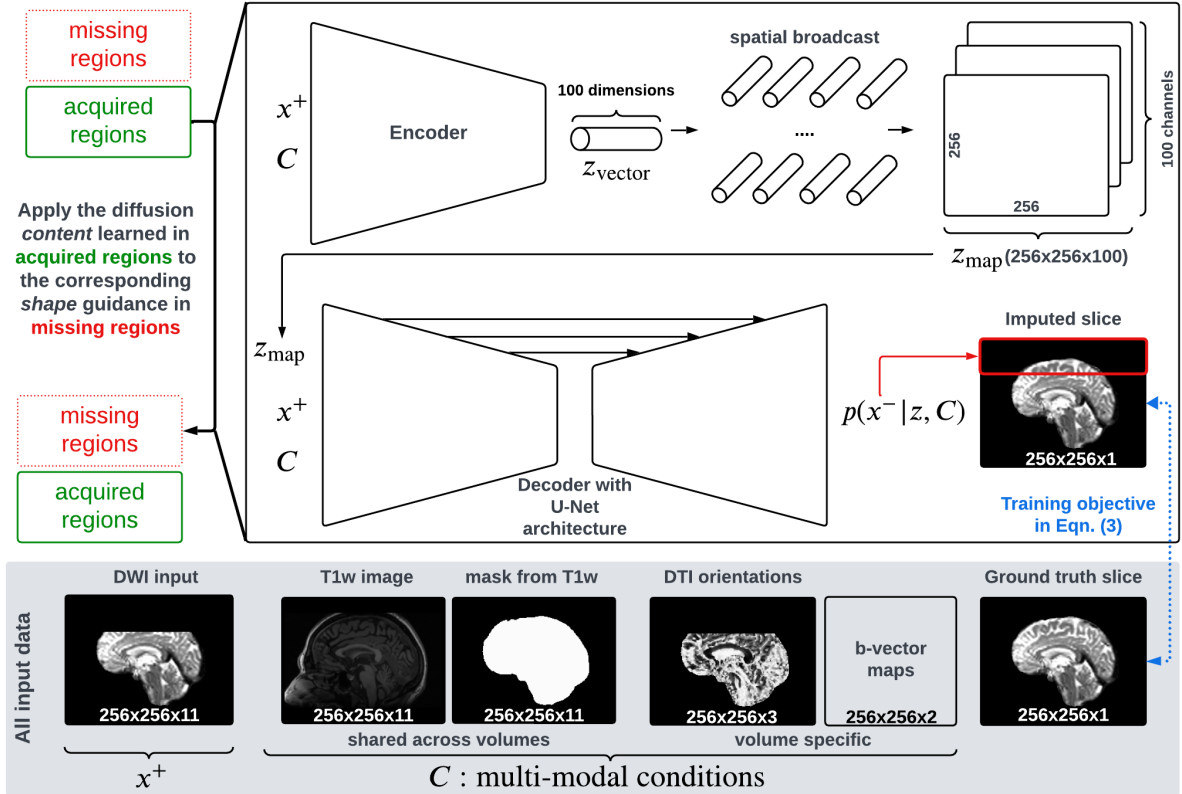

**Figure 4.** The proposed framework tackles the imputation task with two specialized modules. First, a *content* module learns diffusion features in the acquired region using a conditional VAE encoder. Then, the diffusion features are spatially broadcasted to the missing regions, along with the multi-modal conditioning features extracted by a *shape* module. Finally, all information is integrated using a decoder with U-Net architecture to "fill-in" the brain *shape* in the missing regions with the diffusion *content* to impute the missing dMRI slices. The size of each image is presented at the bottom in the order of [height × width × channel].

To enhance the realism of the final generated images and to encourage that the $x^-$ should match well with $x^+$, we additionally apply the generative adversarial network (GAN) objective [40,41] for the whole image as follow:

$$\mathcal{L}_{GAN}(\theta, \phi, D) = \mathbb{E}_y[\log D(y)] + \mathbb{E}_x[\log(1 - D(G_{\theta,\phi}(x, C)))], \quad (2)$$

where $D$ is a discriminator to criticize whether the output of the generative model $G$ looks real. The final objective of our model is then formulated as:

$$\mathcal{L} = \mathcal{L}_{GAN}(\theta, \phi, D) + \mathcal{L}_{VAE}(\theta, \phi) \quad (3)$$

As shown in Figure 4, the input pair of $x^+$ and $C$ are first fed to a VAE encoder to sample a latent feature vector $z$ from its inferred distribution. Next, this vector $z$ is spatially and repeatedly tiled to match the height and width of the images in $C$, and then concatenate with $x^+$ and $C$ as additional image channels. The $x^+$ is introduced here to improve the smoothness of imputation at the boundary between acquired and missing regions, based on our preliminary experiments. At this point, the diffusion features extracted from acquired DWI are spatially aligned with the anatomical structure introduced by the multi-modality images, and then they are fed together into a decoder with U-Net architecture to synthesize a DWI with a complete FOV. The loss defined in Eqn. (3) is then computed between the synthesized DWI and the ground truth images.

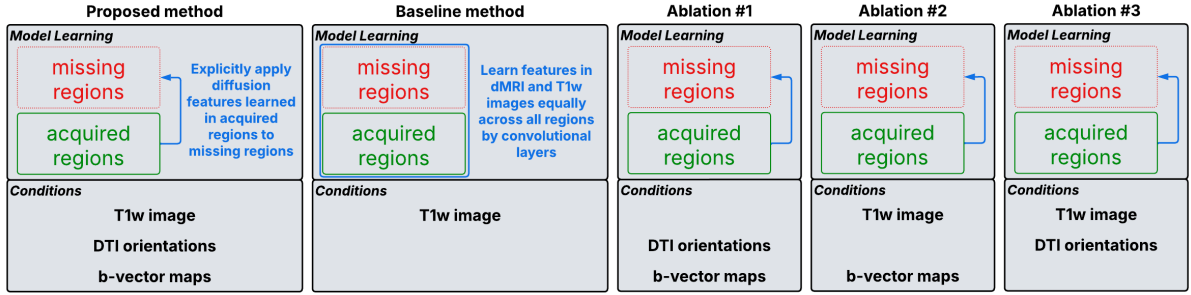

**Figure 5. The key difference among the models studied in this work lies in how they learn features from multiple input images, as well as in their sets of conditional modalities.**

We followed the implementation strategies of our previous work [19]: First, during training we randomly cut off 20 mm to 50 mm from either the top or bottom of the DWI to enhance the robust training of the model. Second, to tackle the large GPU memory requirements associated with learning the 3D networks of our images, we adapt the 2.5D framework of our previous work, where the input of the neural network are patches of the images, where each patch is one sagittal slice accompanied by 5 neighboring slices on both sides, which results in a size of $256 \times 256 \times (2 \times 5 + 1)$. The output is a single imputed slice of DWI corresponding to the input slices, which is later merged to get a complete volume. Third, to address the large difference in image distribution between b0 images and b1300 images, we trained two separate models with the same architecture, each specifically applied to imputing b0 and b1300 images respectively. Model selection is conducted on the validation subjects whose DWI are cut off 30 mm and 50 mm from the top of the brain. For all testing subjects, their DWI are cut off by 50 mm from the top of the brain.

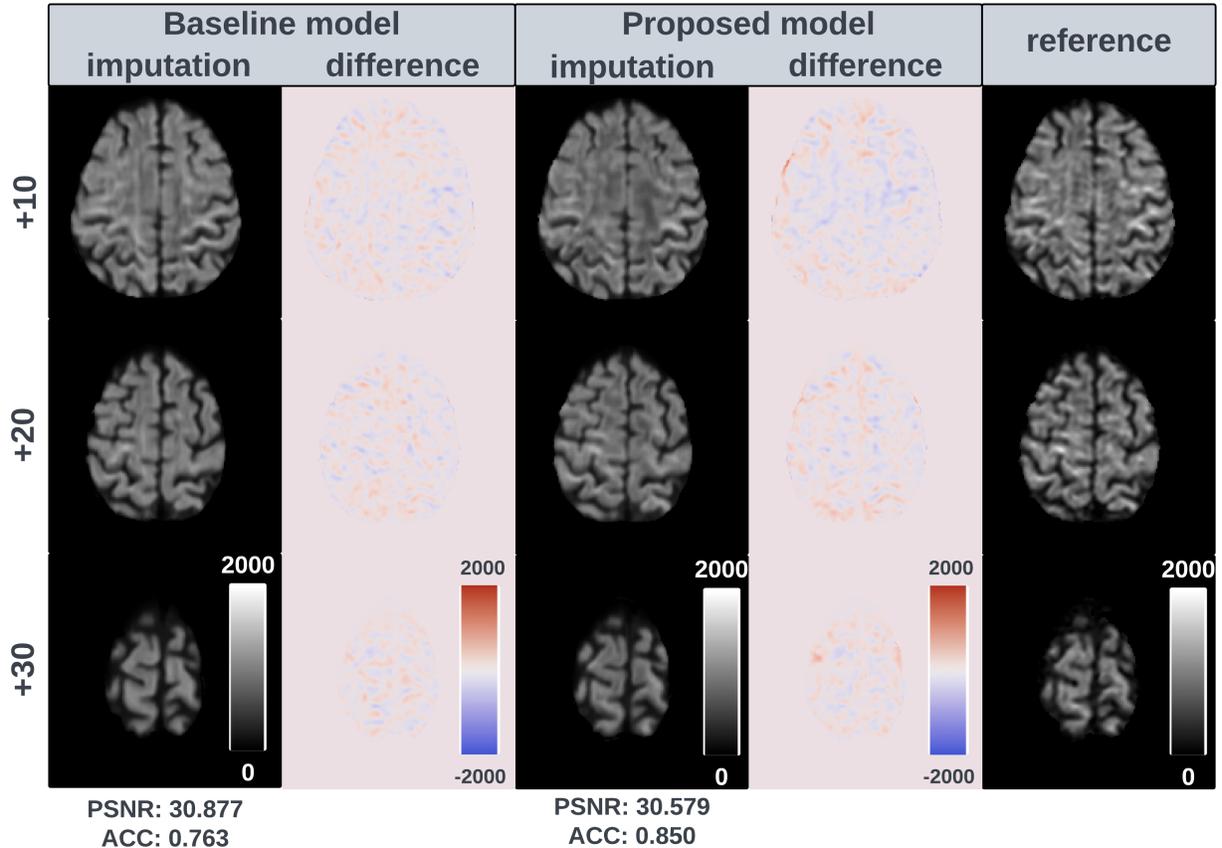

**Figure 6. Axial view of the imputations for b1300 images.** Each row represents a certain distance to the nearest acquired slice in millimeters (mm). Red and blue indicate that the imputed intensity is larger or smaller, respectively, than the ground truth reference. Compared with the baseline model, the proposed model can impute voxel with smaller values (appears darker) for highly structured tissues, which better reflects the MRI signal attenuation due to the diffusion of water molecules. PSNR and ACC values for the presented subject are shown at the bottom of the figure for reference.

*2.4 Analysis*

We conducted the following studies to evaluate the imputation performance of the proposed method and its usefulness in improving the accuracy of whole-brain white matter bundles. To test our hypothesis that the proposed method can address the limitations of existing works for imputing dMRI by specifically integrating distinct information from multi-modality images, we chose our previous work [19] as the baseline, which it also conditioned on T1w images but treated the information equally as the dMRI. We study and compare the proposed methods with baseline on both WRAP and NACC datasets.

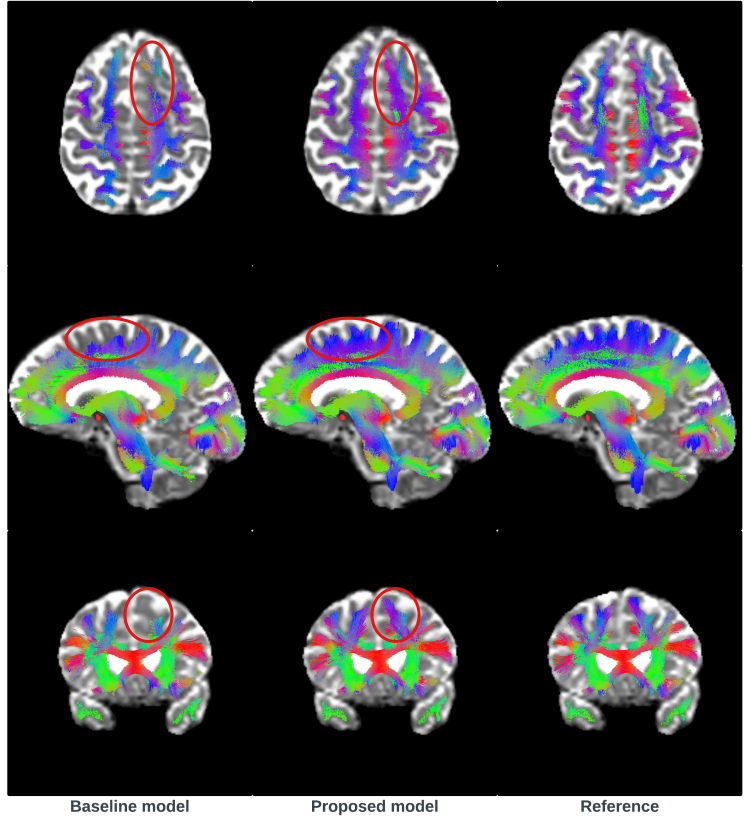

**Figure 7.** Visualization of tracts computed from images imputed by the baseline model compared to the proposed model highlights the more accurate and complete downstream tractography results, as demonstrated by the ability of the proposed model to fix the broken and incomplete streamlines shown in the baseline model. Tractograms are plotted on top of a reference b0 image for anatomical context. Red circles mark some distinguishable regions for demonstration of the improved tractography accuracy of the proposed model.

First, we quantitatively evaluate the error of the imputation. In general, we used SLANT-TICV [42] to compute a brain mask and apply it to the imputed brain to ensure that only brain areas are considered for computing metrics. For the imputed brain regions, we focus on measuring the angular correlation coefficient (ACC) [43], which is widely used for comparisons in dMRI data given its natural directional properties [44–48]. Specifically, we first compute the spherical function [49] for the 4D DWI volumes using the DIPY implementation [50], and then calculate the ACC between the spherical functions of the imputed brain regions and their corresponding ground truth. To carry out analysis of ACC only in fibrous tissue, a subset of SLANT-TICV [42] brain segmentations is further applied, including the right and left cerebellum white matter (label 40 and 41) and the right and left cerebral white matter (label 44 and 45). Additionally, we report the whole brain voxel-wise peak signal-to-noise ratio (PSNR) as a general metric for image imputation. Qualitatively, we present visualizations of the imputed slices of the brain regions in comparison with its ground truth references.

|  |  | **WRAP** |  | **NACC** |  |
|---|---|---|---|---|---|
| **ACC** | baseline | 0.733±0.039 | | 0.703±0.075 | |
|  | proposed | **0.798±0.041** | | **0.731±0.069** | |
|  | p-value | * 1.8E-28 | | * 1.7E-6 | |
|  |  | b0 images | b1300 images | b0 images | b1300 images |
| **PSNR WM** | baseline | **34.30±1.85** | 24.00±1.52 | **32.77±1.93** | **23.21±1.34** |
|  | proposed | 34.21±1.93 | **24.45±1.67** | 32.02±2.32 | 22.81±1.41 |
|  | p-value | 0.029 | * 3.7E-96 | * 5.1E26 | * 5.6E-58 |
| **PSNR Non-WM** | baseline | **28.91±1.83** | **29.24±1.37** | **27.66±2.46** | **29.37±1.26** |
|  | proposed | 28.10±1.66 | 28.97±1.42 | 27.31±2.22 | 28.96±1.27 |
|  | p-value | * 2.1E-9 | * 0.0 | * 3.1E-14 | * 0.0 |

**Table 1.** Average ACC and PSNR for the imputation regions of testing data on WRAP and NACC datasets. The proposed method achieved significant improvement for all-volume-wise imputation, as demonstrated by the ACC values. No significant difference is observed when imputing both white matter (WM) and non-WM regions, as demonstrated by the voxel-wise based metric, PSNR.

Next, to test our hypothesis that an improved imputation of the incomplete part of FOV can improve the whole brain tractography, we conduct whole brain bundle analysis and evaluate the tracts produced from images imputed by the proposed methods and the baseline methods. To ensure an accurate comparison, we extracted tracts using Tractseg [51] from images imputed by both the baseline and proposed methods. These tracts were then compared against the same tract segmentation derived from the ground truth reference image with a complete FOV, and the Dice score was calculated for each imputation method. We conduct paired t-tests for the 72 tracts and present the visualization of the streamlines produced from different imputation methods and mark the difference and improvement compared with ground truth reference streamlines. Specifically, we investigate a group of 10 tracts that are commonly associated with neurodegenerative including Alzheimer's disease and general aging [52–66], which are Rostrum (CC_1) and Genu (CC_2) of the Corpus Callosum (CC) as well as left and right Cingulum (CG), Fornix (FX), Inferior occipito-frontal fascicle (IFO), and Superior longitudinal fascicle I (SLF_I), for exploring potential clinical benefits of the proposed framework. We present Bland-Altman plots to analyze the agreement in the shape measurements of these bundles between the reference and various imputation methods.

|  | WRAP | | NACC | |
| --- | --- | --- | --- | --- |
|  | AD-associated bundles | All bundles | AD-associated bundles | All bundles |
| *Baseline* | 0.838±0.049 | 0.844±0.052 | 0.821±0.066 | 0.839±0.060 |
| *Proposed method* | **0.858±0.039** | **0.865±0.038** | **0.839±0.054** | **0.854±0.046** |
| *p-value* | * 0.0061 | * 2.0E-13 | * 0.0039 | * 2.7E-8 |

Table 2. Average Dice score for 72 tracts produced from the imputed images. The improvement of the proposed model over the baseline model is statistical significant (all p-values is smaller than 0.01) from paired t-test conducted for all-bundles and bundles associated with AD.

Finally, to evaluate the contribution and effect of each input conditional modality, we conducted ablation studies by systematically removing each conditional modality from the proposed method: the T1w image, DTI orientation image, and b-vector maps. Figure 5 presents a visualization of the model comparisons, including the ablation models and the baseline model.

## 3 Results

### 3.1 Imputation of the incomplete part of the FOV

The proposed model, in comparison to the baseline model, can more accurately impute smaller voxel values (resulting in a darker appearance) for highly structured tissues in white matter (Figure 6). This improvement better represents the dMRI signal attenuation associated with water molecule diffusion along the directions of nerve fibers. This difference in imputation is clearly demonstrated by the volumetric metric, ACC, where the proposed method achieved relative improvements of 8.9% and 4.0% on the WRAP and NACC datasets, respectively. These improvements are statistically significant, as indicated by the small p-value (Table 1 top). The substantial improvement in ACC shows that the imputed regions generated by the proposed method preserve the angular structure, which is critical for downstream tasks such as tractography and fiber orientation estimation. Regarding the voxel-wise measurement, PSNR, the difference between the baseline method and the proposed method is marginal in both white-matter regions and non-white-matter regions (Table 1 bottom). Given the 4D nature of dMRI, PSNR may underrepresent diffusion structural fidelity, especially when considering directional agreement between the imputation and its ground truth. In conclusion, while PSNR improvements are marginal due to its voxel-wise nature of measurement, the substantial gains in ACC underscore the effectiveness of the proposed method in preserving directional diffusion signals. Given the 4D directional structure of dMRI and the importance of orientation accuracy for downstream tasks, ACC offers a more meaningful evaluation of imputation quality in dMRI.

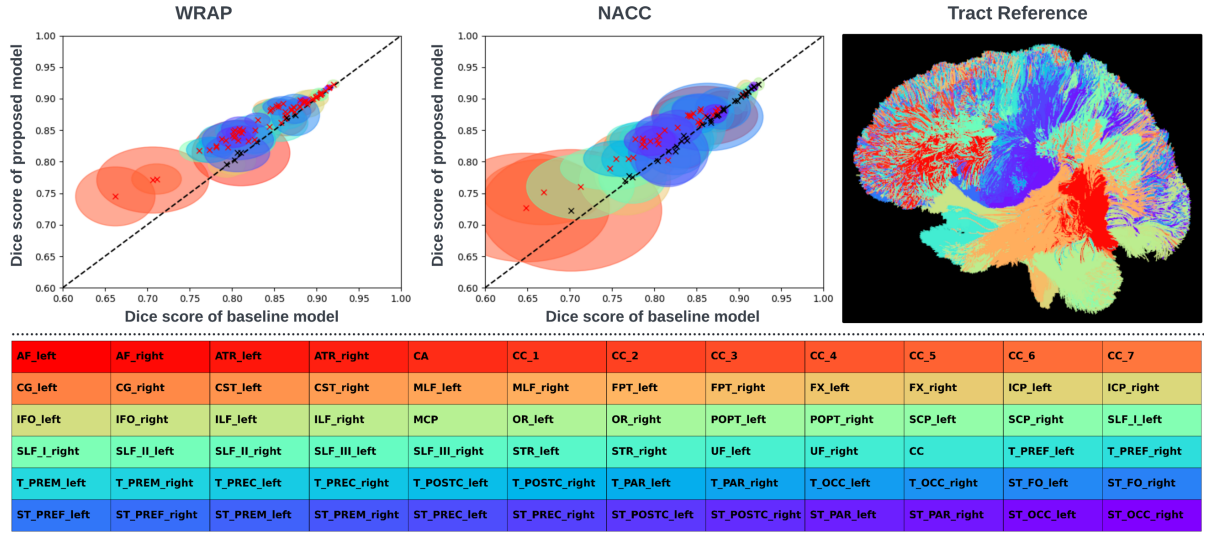

**Figure 8.** Comparison of Dice scores for the baseline model and the proposed model. In the top left two figures, each colored ellipse represents one tract. The center of each ellipse is determined by the Dice scores of the baseline model (x-coordinate) and the proposed model (y-coordinate). The horizontal and vertical axes of the ellipse represent the standard deviation (STD) of the Dice scores for the baseline and proposed models, respectively. The corresponding reference for each tract is plotted in the top right figure, where the colors of tracts match that in the left two figures. The legends for tract colors are presented in the bottom figure. The proposed model consistently shows improvement for almost all tracts, as indicated by all ellipses are located above the black dashed line (y=x), which represents equal performance. For tracts where the difference is significant (p<0.01), the centers are colored red; otherwise, the centers are black.

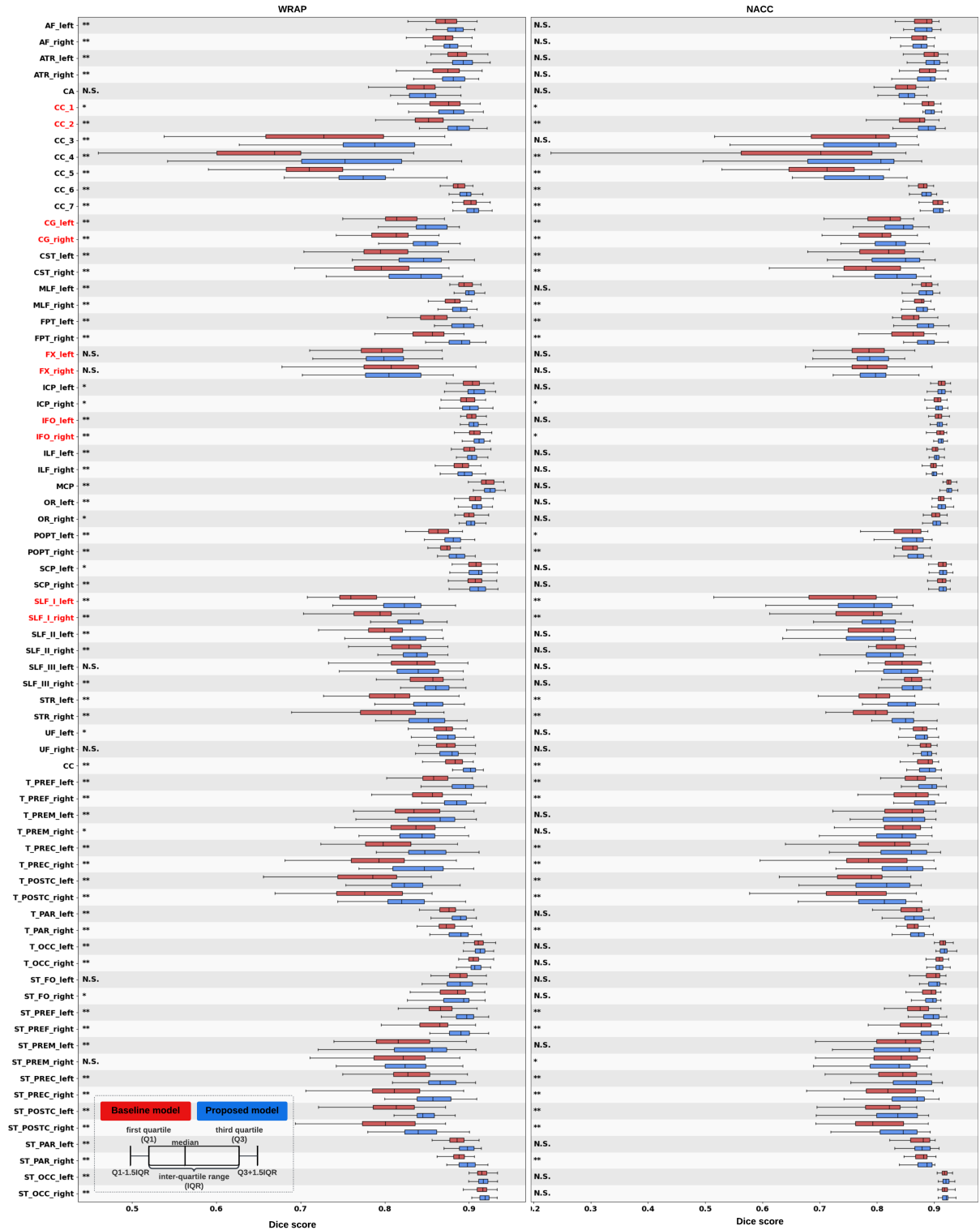

**Figure 9.** In both the WRAP and NACC datasets, the proposed framework (blue) improves tractography accuracy by imputing the incomplete part of the FOV, as demonstrated by the overall higher Dice scores compared to the baseline model (red). Tracts commonly associated with neurodegenerative have their names in red. Statistical significance for each tract is denoted by "*" (p < 0.05), "**" (p < 0.01), and "N.S." (not significant).

*3.2 White matter bundle analysis*

The proposed method achieved more complete and more accurate tractography compared with the baseline model, especially in regions close to the top edges of the brain (Figure 7). The streamlines computed by the baseline model are either broken or terminate progression early toward the top of the brain. However, the proposed method addresses this issue by enabling the streamlines to propagate further and extend over a longer range, thus enhancing the visualization and analysis of neural pathways in whole brain areas. The quantitative results further support the improved imputation achieved by the proposed method, demonstrated by consistent enhancements in average bundle segmentation across all bundles, including those associated with Alzheimer's Disease (Table 2). All improvements are statistically significant, with p-values less than 0.01, highlighting the robustness of the proposed method in imputing various neural pathways. Additionally, we conduct a detailed comparison of Dice scores achieved by the baseline and the proposed models for the segmentation map of each individual tract. Notably, the proposed model consistently achieves improvement for almost all tracts, as evidenced by the ellipses of every tract predominantly positioned above the black dashed line, which denotes equal performance between the baseline and the proposed method (Figure 8). The improvements are statistically significant for most tracts ($p<0.01$, highlighted by red centers) on both the WRAP and NACC datasets. The corresponding reference for each tract is presented in the right figure, where the color coding matches that of the ellipses, aiming for a clear visual correlation and aiding in the rapid identification of each tract's accuracy metrics and its position in the brain. Additionally, full visualization comparisons for all 72 tracts confirm that the proposed method consistently achieves higher Dice scores compared to the baseline model (Figure 9). Finally, Bland-Altman plots for examining the shape measurements of the 12 AD associated bundles are presented in Figure 10. Our approach shows significantly more consistent agreement with the reference measurements than the measurements obtained from images imputed by the baseline method, thereby reducing the uncertainty in analyzing bundles associated with Alzheimer's Disease and allowing better recovery of whole brain tractography from corrupted data.

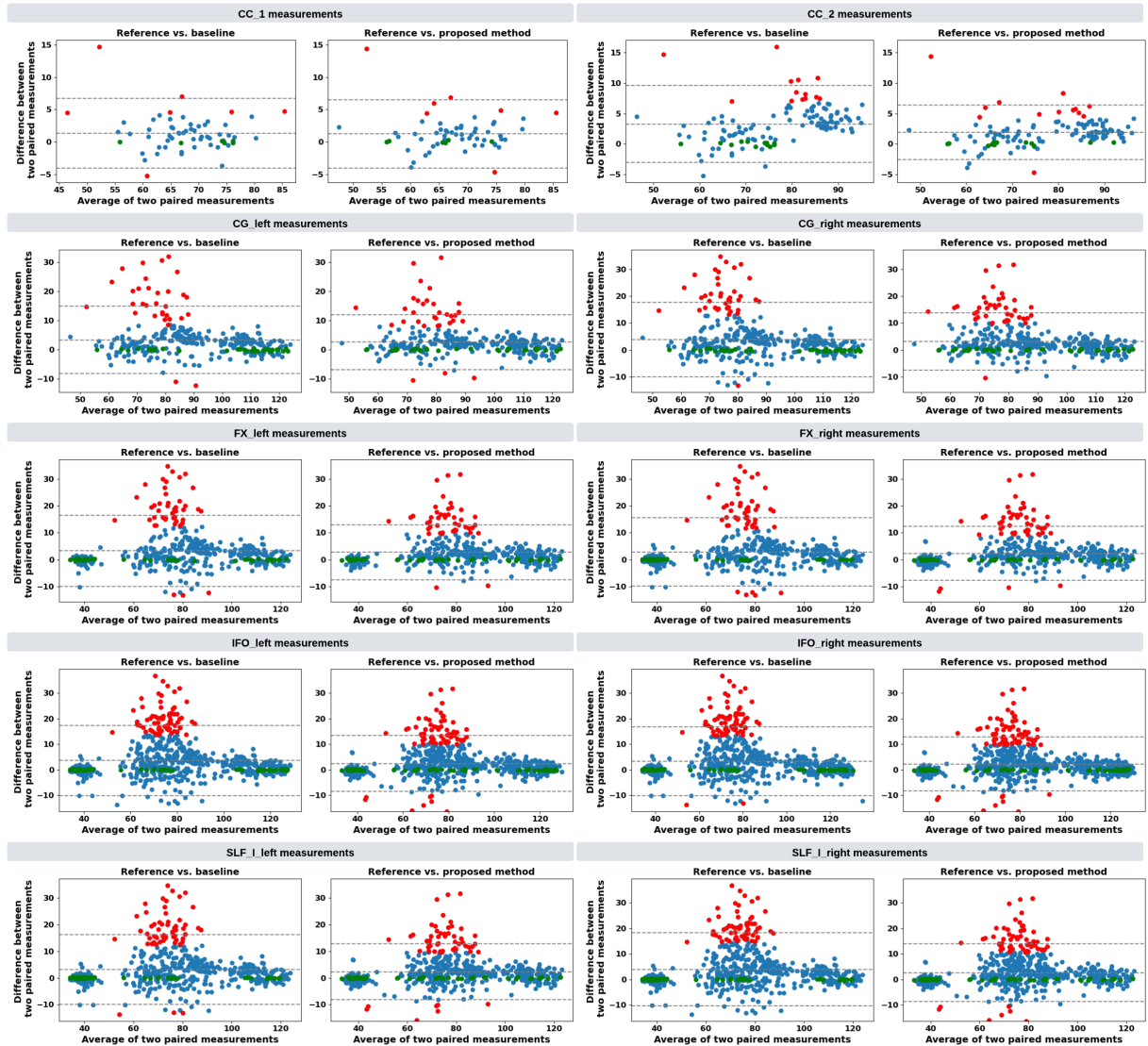

**Figure 10. Bland-Altman plots illustrating the agreement on average bundle length compared to the reference. Measurements within the best 10% for accuracy are marked in green, while those in the worst 10% are in red. The analysis specifically investigates the tracts associated with neurodegenerative including AD such as CC_1, CC_2, CG, FX, IFO, and SLF_I. Our method reduces the major variations in measurements caused by incomplete FOV, compared to the baseline method. In the "Reference vs. proposed method" plots, measurements are tightly clustered near the middle dashed line, demonstrating consistent and coherent agreement with the reference. Conversely, the "Reference vs. baseline" plots reveal a wider distribution of measurements along the y-axis, indicating considerable errors and variability. By ensuring consistent measurements of neurodegenerative-associated bundles, our approach decreases the uncertainty in neurodegenerative studies potentially compromised by incomplete FOV.**

| Dataset | Proposed method | No T1w image | No DTI orientations | No b-vector maps |
|---|---|---|---|---|
| WRAP | **0.798±0.041** | 0.296±0.040 | 0.535±0.054 | 0.505±0.050 |
| NACC | **0.731±0.069** | 0.247±0.042 | 0.425±0.058 | 0.405±0.052 |

**Table 3. Removing any of conditional inputs leads to a substantial decline in ACC.**

*3.3 Ablation Study*

All conditional modalities provide important information as input to the proposed method, as demonstrated by the substantial decline in ACC observed after removing any of them, as measured on both the WRAP and NACC datasets (Table 3). Specifically, excluding the T1w image results in the most significant performance drop (WRAP: 63%, NACC: 66%), highlighting its critical role in the anatomical structure guidance. Omitting DTI orientation images (WRAP: 33% drop, NACC: 41% drop), or b-vector maps (WRAP: 37% drop, NACC: 45% drop) also leads to notable worse performance. The proposed method, when using all conditional modalities, achieves the highest performance.

**4 Discussion**

In this study, we introduced a multi-modality conditioned variational U-net, aimed at addressing the limitations of existing methods for imputing incomplete FOV in dMRI. Our framework demonstrates substantial improvements over the baseline model, particularly in its capacity to integrate diverse imaging modalities to enhance imputation and subsequent whole-brain tractography.

**Integration of multi-modality information:** One of the standout features of our approach is the effective and explicit integration of information from multi-modality images for dMRI FOV extension. By spatially broadcasting the diffusion features learned in acquired dMRI to the anatomical structure guided by the multi-modality images, our approach effectively "fill-in" the known diffusion *content* to the brain *shape* of the regions that require imputation. This novel design of our method achieved more accurate volumetric imputation that better reflect the signal attenuation in dMRI due to the white matter tissue properties, which is supported by both qualitative visualization (Figure 6) and the quantitative volumetric metric, ACC (Table 1). These results and the ablation study (Table 3) support our hypothesis that by explicitly integrating distinct information from multi-modality images the proposed method can improve the imputation of the incomplete FOV in dMRI.

**Enhancements in Tractography:** The proposed model significantly improves whole brain tractography, especially addressing the limitations that the baseline method fails to overcome, such as producing streamlines near the top edge of the brain. These evaluations and findings on 72 tracts support our hypothesis that an improved imputation of the incomplete part of FOV can improve the whole brain tractography. Furthermore, accurate whole brain tractography is essential for understanding the connectivity of neural pathways that may be affected by neurodegenerative diseases. By improving the completeness and accuracy of whole brain tractography our method provides a more reliable technique for potential clinical studies and interventions.

**Applications for Neurodegenerative Studies:** For neurodegenerative disorders like Alzheimer's Disease, where early detection and accurate monitoring of disease progression are crucial, the ability to accurately reconstruct and analyze brain tracts can significantly influence therapeutic strategies and outcomes, especially when dealing with dMRI data that may contain incomplete FOV. By providing a more reliable imputation technique, our method is useful for repairing incomplete data and further aids in a deeper understanding of neurodegenerative and its impact on brain connectivity. Specifically, the reliability of our approach is demonstrated through Bland-Altman plots for tracts associated with neurodegenerative including AD, which revealed a more consistent agreement with reference tract measurements, highlighting the potential of our approach to reduce diagnostic uncertainties and enhance monitoring of disease progression by fixing valuable dMRI data.

**Limitation:** The successful imputation results achieved in this work depend on a key assumption that the diffusion characteristics are shared between the incomplete FOV regions and the acquired regions. This assumption may break down when encountering images with pathological or abnormal conditions, where simply applying the same diffusion characteristics from one region to another can be problematic. This may lead to inaccurate or biased imputations in regions affected by lesions, tumors, or other abnormalities, potentially limiting the clinical applicability of the method. Future work could explore incorporating pathology-aware priors that account for local variations in diffusion characteristics.

## 5 Conclusion

This study successfully demonstrates that the proposed multi-modality conditioned variational U-net framework significantly enhances the imputation for the incomplete part of FOV in dMRI scans. By specifically incorporating diffusion features in acquired dMRI along with distinct anatomical structure information in multi-modality conditions, our framework not only improves the imputation performance of missing brain regions but also facilitates more accurate and comprehensive whole-brain tractography. The substantial improvements in angular correlation coefficients and Dice scores, compared to the baseline method, demonstrate the effectiveness of our approach in leveraging multi-modality information to enhance the quality of dMRI imputations. These enhancements are useful for advancing the reliability of neuroimaging analyses with potentially corrupted data with incomplete FOV, particularly in the context of neurodegenerative diseases like Alzheimer's Disease, where precise whole brain mapping of neural pathways is essential for understanding disease progression. In future work, we aim to further refine the models that are capable of more effective integration of modalities within more explicitly anatomical structure prior and explore its application to further enhance tractography analysis. This will potentially expand the utility of our approach, ensuring that it can meet a broad range of clinical and research needs, paving the way for more sophisticated and precise neuroimaging studies for critical issues like Alzheimer's Disease.

**CRediT authorship contribution statement**

**Zhiyuan Li:** Conceptualization, Methodology, Software, Validation, Formal analysis, Investigation, Data Curation, Writing – original draft, Visualization, Project administration. **Chenyu Gao:** Methodology, Writing - Review & Editing. **Praitayini Kanakaraj:** Methodology, Writing - Review & Editing. **Shunxing Bao:** Methodology, Writing - Review & Editing. **Lianrui Zuo:** Methodology, Writing - Review & Editing. **Michael E. Kim:** Data Curation, Writing - Review & Editing. **Nancy R. Newlin:** Data Curation, Writing - Review & Editing. **Gaurav Rudravaram:** Data Curation, Writing - Review & Editing. **Nazirah M. Khairi:** Data Curation,

Writing - Review & Editing. **Yuankai Huo:** Resources, Writing - Review & Editing. **Kurt G. Schilling:** Resources, Writing - Review & Editing. **Walter A. Kukull:** Resources, Data Curation, Writing - Review & Editing. **Arthur W. Toga:** Resources, Data Curation, Funding acquisition, Writing - Review & Editing. **Derek B. Archer:** Resources, Funding acquisition, Writing - Review & Editing. **Timothy J. Hohman:** Resources, Data Curation, Funding acquisition, Writing - Review & Editing. **Bennett A. Landman:** Supervision, Methodology, Conceptualization, Writing - Review & Editing, Project administration, Funding acquisition.

**Data availability**

The data were used under agreement for this study and are therefore not publicly available. More information about the datasets can be found at NACC (https://www.naccdata.org/) and WRAP (https://wrap.wisc.edu/).

**Declaration of competing interest**

None.

**Declaration of generative AI and AI-assisted technologies in the writing process**

During the preparation of this work the author(s) used ChapGPT-4o in order to check the grammar and polish the language. After using this tool/service, the author(s) reviewed and edited the content as needed and take(s) full responsibility for the content of the publication.

**Funding**

This research is supported by NSF CAREER 1452485, NIH 1R01EB017230, NIH NIDDK K01-EB032898, NIA K01-AG073584, and NIA U24AG074855. This study was supported in part using the resources of the Advanced Computing Center for Research and Education (ACCRE) at Vanderbilt University, Nashville, TN (NIH S10OD023680). We gratefully acknowledge the support of NVIDIA Corporation with the donation of the Quadro RTX 5000 GPU used for this research. The imaging dataset(s) used for this research were obtained with the support of ImageVU, a research resource supported by the Vanderbilt Institute for Clinical and Translational Research (VICTR) and Vanderbilt University Medical Center institutional funding. The VICTR is funded by the National Center for Advancing Translational Sciences (NCATS) Clinical Translational Science Award (CTSA) Program, Award Number 5UL1TR002243-03. The content is solely the responsibility of the authors and does not necessarily represent the official views of the NIH. The ADSP Phenotype Harmonization Consortium (ADSP-PHC) is funded by NIA (U24 AG074855, U01 AG068057 and R01 AG059716). The harmonized cohorts within the ADSP-PHC included in this manuscript were: the National Alzheimer's Coordinating Center (NACC): The NACC database is funded by NIA/NIH Grant U24 AG072122. NACC data are contributed by the NIA-funded ADRCs: P30 AG062429 (PI James Brewer, MD, PhD), P30 AG066468 (PI Oscar Lopez, MD), P30 AG062421 (PI Bradley Hyman, MD, PhD), P30 AG066509 (PI Thomas Grabowski, MD), P30 AG066514 (PI Mary Sano, PhD), P30 AG066530 (PI Helena Chui, MD), P30 AG066507 (PI Marilyn Albert, PhD), P30 AG066444 (PI John Morris, MD), P30 AG066518 (PI Jeffrey Kaye, MD), P30 AG066512 (PI Thomas Wisniewski, MD), P30 AG066462 (PI Scott Small, MD), P30 AG072979 (PI David Wolk, MD), P30 AG072972 (PI Charles DeCarli, MD), P30 AG072976 (PI Andrew Saykin, PsyD), P30 AG072975 (PI David Bennett, MD), P30 AG072978 (PI Neil Kowall, MD), P30 AG072977 (PI Robert Vassar, PhD), P30 AG066519 (PI Frank LaFerla, PhD), P30 AG062677 (PI Ronald Petersen, MD, PhD), P30 AG079280 (PI Eric Reiman, MD), P30 AG062422 (PI Gil Rabinovici, MD), P30 AG066511 (PI Allan Levey, MD, PhD), P30 AG072946 (PI Linda Van Eldik, PhD), P30 AG062715 (PI Sanjay Asthana, MD, FRCP), P30

AG072973 (PI Russell Swerdlow, MD), P30 AG066506 (PI Todd Golde, MD, PhD), P30 AG066508 (PI Stephen Strittmatter, MD, PhD), P30 AG066515 (PI Victor Henderson, MD, MS), P30 AG072947 (PI Suzanne Craft, PhD), P30 AG072931 (PI Henry Paulson, MD, PhD), P30 AG066546 (PI Sudha Seshadri, MD), P20 AG068024 (PI Erik Roberson, MD, PhD), P20 AG068053 (PI Justin Miller, PhD), P20 AG068077 (PI Gary Rosenberg, MD), P20 AG068082 (PI Angela Jefferson, PhD), P30 AG072958 (PI Heather Whitson, MD), P30 AG072959 (PI James Leverenz, MD); National Institute on Aging Alzheimer's Disease Family Based Study (NIA-AD FBS): U24 AG056270; Religious Orders Study (ROS): P30AG10161,R01AG15819, R01AG42210; Memory and Aging Project (MAP - Rush): R01AG017917, R01AG42210; Minority Aging Research Study (MARS): R01AG22018, R01AG42210; Washington Heights/Inwood Columbia Aging Project (WHICAP): RF1 AG054023; and Wisconsin Registry for Alzheimer's Prevention (WRAP): R01AG027161 and R01AG054047. Additional acknowledgments include the National Institute on Aging Genetics of Alzheimer's Disease Data Storage Site (NIAGADS, U24AG041689) at the University of Pennsylvania, funded by NIA.